\documentclass[conference]{IEEEtran}
\IEEEoverridecommandlockouts
\usepackage{cite}
\usepackage{amsmath,amssymb,amsfonts}
\usepackage{algorithmic}
\usepackage{graphicx}
\usepackage{textcomp}
\usepackage{xcolor}
\usepackage{booktabs}
\usepackage{subcaption}
\usepackage{multicol}
\def\BibTeX{{\rm B\kern-.05em{\sc i\kern-.025em b}\kern-.08em
    T\kern-.1667em\lower.7ex\hbox{E}\kern-.125emX}}
\begin{document}

\title{Manipulating Soft Tissues by Deep Reinforcement Learning for Autonomous Robotic Surgery
%\thanks{This paper is an extended abstract version. The full paper will be submitted once acceptance.}
}

\author{
\IEEEauthorblockN{Ngoc Duy Nguyen}
\IEEEauthorblockA{\textit{Institute for Intelligent Systems}\\ \textit{Research and Innovation (IISRI)} \\
\textit{Deakin University}\\
Geelong, Victoria, Australia \\
duy.nguyen@deakin.edu.au}
\and
\IEEEauthorblockN{Thanh Nguyen}
\IEEEauthorblockA{\textit{Institute for Intelligent Systems}\\ \textit{Research and Innovation (IISRI)} \\
\textit{Deakin University}\\
Geelong, Victoria, Australia \\
thanh.nguyen@deakin.edu.au}
\and
\IEEEauthorblockN{Saeid Nahavandi}
\IEEEauthorblockA{\textit{Institute for Intelligent Systems}\\ \textit{Research and Innovation (IISRI)} \\
\textit{Deakin University}\\
Geelong, Victoria, Australia \\
saeid.nahavandi@deakin.edu.au}
\and
\IEEEauthorblockN{Asim Bhatti}
\IEEEauthorblockA{\textit{Institute for Intelligent Systems Research and Innovation (IISRI)} \\
\textit{Deakin University}\\
Geelong, Victoria, Australia \\
asim.bhatti@deakin.edu.au}
\and
\IEEEauthorblockN{Glenn Guest}
\IEEEauthorblockA{\textit{Faculty of Health}\\
\textit{Deakin University}\\
Geelong, Victoria, Australia \\
g.guest@deakin.edu.au}
}

\maketitle

\begin{abstract}

In robotic surgery, pattern cutting through a deformable material is a challenging research field. The cutting procedure requires a robot to concurrently manipulate a scissor and a gripper to cut through a predefined contour trajectory on the deformable sheet. The gripper ensures the cutting accuracy by nailing a point on the sheet and continuously tensioning the pinch point to different directions while the scissor is in action. The goal is to find a pinch point and a corresponding tensioning policy to minimize damage to the material and increase cutting accuracy measured by the symmetric difference between the predefined contour and the cut contour. Previous study considers finding one fixed pinch point during the course of cutting, which is inaccurate and unsafe when the contour trajectory is complex. In this paper, we examine the soft tissue cutting task by using multiple pinch points, which imitates human operations while cutting. This approach, however, does not require the use of a multi-gripper robot. We use a deep reinforcement learning algorithm to find an optimal tensioning policy of a pinch point. Simulation results show that the multi-point approach outperforms the state-of-the-art method in soft pattern cutting task with respect to both accuracy and reliability.

\end{abstract}


\section{Introduction}
\begin{figure}[!t]
\centering
\includegraphics[width=0.68\linewidth]{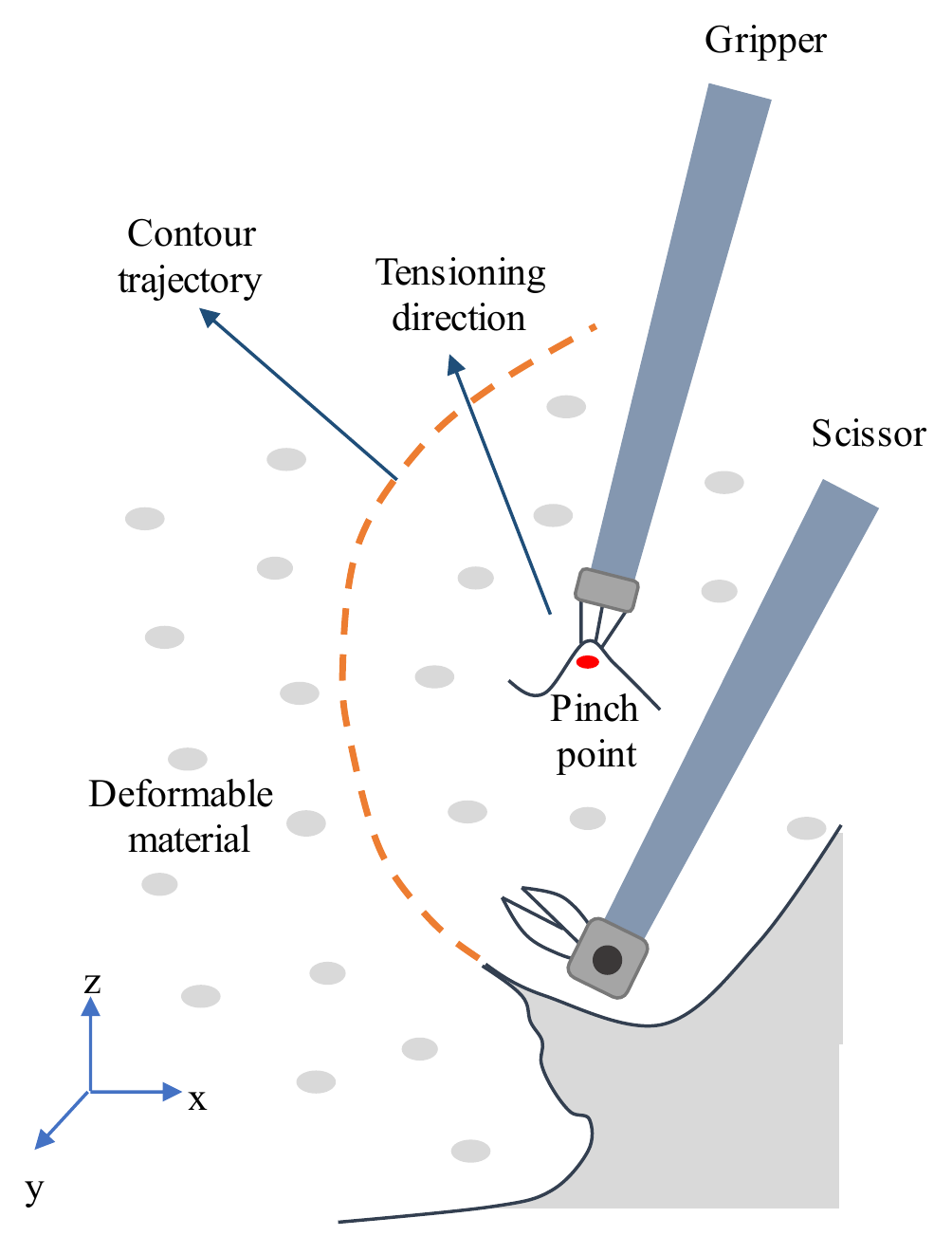}
\caption{A surgical pattern cutting task.}
\label{fig:1} 
\end{figure}

In robotic surgery, manipulation of a deformable sheet, especially cutting through a predefined contour trajectory, is a critical task that has attracted a significant number of research interests \cite{01, 02, 03, 04, 04a}. The pattern cutting task is one of the \emph{Fundamental Skills of Robotic Surgery} (FSRS) because it minimizes surgeon errors, operation time, trauma, and expenses \cite{11, 12, 13}. Furthermore, the deformable material is usually soft and elastic, which is intractable to perform a cutting procedure accurately \cite{05}. Therefore, it is necessary to use a \emph{gripper} \cite{06, 07}, which holds a point (\emph{pinch point}) on the sheet and tensions it along an allowable set of directions with a reasonable force while a surgical \emph{scissor} is used to cut through a \emph{contour trajectory}, as shown in Fig.~\ref{fig:1}. 

In other words, the pattern cutting task involves two essential steps: 1) selecting a pinch point and 2) finding a tensioning policy from that pinch point. Previous study considers a single pinch point over the course of cutting, which is only efficient when the contour shape is simple. Conversely, it is more appropriate to divide a complicated contour into different segments. In this case, the use of one pinch point is unsafe and significantly reduces cutting accuracy \cite{09}. 

In this paper, we examine a multi-point approach and compare the accuracy with its counterpart. Because the robot has a single gripper, only one pinch point is used for tensioning and the others are pinned permanently in the setup phase. Finally, we use a deep reinforcement learning algorithm, namely \emph{Trust Region Policy Optimization} (TRPO) \cite{010}, as in \cite{09}, to seek the optimal tensioning policy from a pinch point. The tensioning policy determines the tensioning direction based on the current state of the sheet, contour information, and cutting position.

In this study, we use the simulator described in \cite{09} to evaluate the accuracy and reliability of the proposed approach and compare its performance with the state-of-the-art method \cite{09}. For the sake of conciseness, we call the method described in \cite{09} the \emph{Single-point Deep Reinforcement Learning Tensioning} (SDRLT) method. Furthermore, the simulator has a practical perspective because the learned tensioning policy is reevaluated by the well-known physical surgical system, \emph{da Vinci Research Kit} (dVRK) \cite{010b}. Simulation results show that the multi-point approach outperforms the SDRLT method and achieves an average of 55\% better accuracy than the non-tensioned baseline over a set of 14 multi-segment contours.

Finally, the paper has the following contributions:

\begin{enumerate}
\item This work provides the first study of using multiple pinch points in pattern cutting task. The study shows the benefits of using multiple pinch points, particularly in complicated contours where a scissor is instructed to cut multiple segments to complete the whole contour trajectory.

\item The proposed scheme outperforms the state-of-the-art method in pattern cutting task and becomes a premise to develop a significant number of research extensions such as the use of multiple grippers, multiple scissors, and multi-layer pattern cutting in a 3D environment.

\item The multi-point approach imitates human demonstrations while cutting. Therefore, the  proposed scheme is useful in both practical implication and theoretical analysis. Finally, the proposed scheme achieves high accuracy and reliability in pattern cutting task.
\end{enumerate}

The rest of the paper is organized as follows. The next section reviews recent advances in surgical automation and the benefits of using reinforcement learning in surgical tasks. Section \ref{sec:3} presents a preliminary background of pattern cutting task and introduces our proposed scheme. Section \ref{sec:4} discusses the experimental results of the proposed scheme and Section \ref{sec:5} concludes the paper.

\section{Related Work}

The use of robot-assisted surgery allows doctors to automate a complicated task with minimal errors and effort. A number of automation levels have been discussed extensively in the literature \cite{14, 15, 16, 17, 17b, 18, 19}. Specifically, early approaches required the existence of experts to create a model trajectory that is used to teach a robot to automatically complete a designated task. For example, Schulman \emph{et al.} \cite{20} use a trajectory transfer algorithm to transform human demonstrations into model trajectories. These trajectories are updated to adapt to new environment geometry and hence assist a robot in learning the task of suturing. Recently, Osa \emph{et al.} \cite{21} propose a framework for online trajectory planning. The framework uses a statistical method to model the distribution of demonstrated trajectories. As a result, it is possible to infer the trajectories into a dynamic environment based on the conditional distribution.

Reinforcement learning (RL) has become a promising approach to modeling an autonomous agent \cite{210, 21a, 21b, 21c, 21d}. RL has the abilities to mimic human learning behaviors to maximize the long-term reward. As a result, RL enables a robot to learn on its own and partially eliminates the existence of experts. Examples of these agents are the box-pushing robot \cite{22}, pole-balancing humanoid \cite{23}, helicopter control \cite{24}, soccer-playing agent \cite{25}, and table tennis playing agent \cite{26}. Furthermore, RL has been utilized in a significant number of research interests in surgical tasks \cite{27, 28}. For example, Chen \emph{et al.} \cite{29} propose a data-driven workflow that combines \emph{Programming by Demonstration} \cite{30} with RL. The workflow encodes the inverse kinematics using trajectories from human demonstrations. Finally, RL is used to minimize the noise and adapt the learned inverse kinematics to the online environment.

Recent breakthrough \cite{31} combines neural networks with RL (deep RL), which enables traditional RL methods to work properly in high-dimensional environments. Typically, Thananjeyan \emph{et al.} \cite{09} use a deep RL algorithm (TRPO) to learn the tensioning policy from a pinch point in pattern cutting task. The tensioning problem is described as a \emph{Markov Decision Process} \cite{32} where each action moves the gripper 1mm along one of four directions in the 2D space. A state of the environment is a state of the deformable sheet, which is represented by a rectangular mesh of point masses. The goal is to minimize the symmetric difference between the ideal contour and the achieved contour cut. This approach, however, uses a single pinch point to complete the pattern cutting task. In this paper, we examine the use of multiple pinch points over the course of cutting. Finally, we compare the cutting accuracy of our proposed scheme with two baseline methods described in \cite{09}: 1) the non-tensioned scheme and 2) SDRLT. Section \ref{sec:3} and Section \ref{sec:4} present the proposed scheme in more details.

\section{Proposed Scheme}
\label{sec:3}

\subsection{Preliminary}
\label{sec:3.1}

\begin{figure}[!t]
\centering
\includegraphics[width=1.\linewidth]{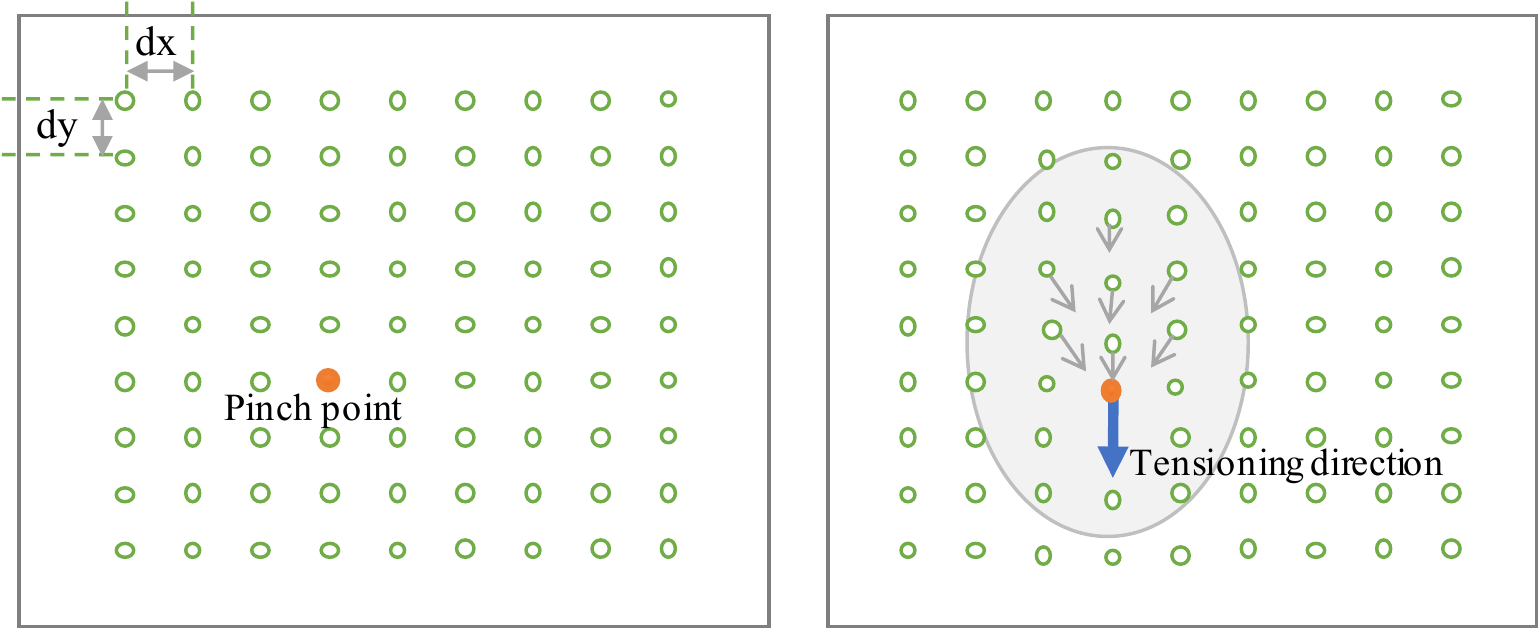}
\caption{A deformable sheet is represented by a mesh of point masses. The left figure illustrates a pinch point without tensioning. The right figure illustrates a pinch point that is tensioned along vertical direction.}
\label{fig:2} 
\end{figure} 

As mentioned earlier, the tensioning problem is described as a Markov Decision Process where a state of the environment is represented by a mesh of $N$ point masses. Initially, these points are aligned evenly in the horizontal direction by a distance $dx$ and in the vertical direction by a distance $dy$ (in the 2D space), as illustrated in Fig. \ref{fig:2}. A set of $N$ points is indexed by $\Sigma = \{1, 2, ..., N\}$. We define $L_\Sigma = \{p_i | p_i \in \mathbf{R}^3,  i \in \Sigma \}$, where $p_i$ denotes a position of point $i$ in 3D space. Initially, we assume that $p_i |_z = 0, \forall i \in \Sigma$. Therefore, a state of the environment at time $t$ is represented by $L^t_\Sigma$.

To simulate the deformable material properly, each point $i$ is connected to its neighbors $\Delta_N^i$ by a spring force. This force maintains the distance between neighboring points. A point is moved to a cut set $\Delta_C$ if it does not have any constraints with its neighbors, \emph{i.e.}, the point is cut by a scissor. When we apply an external force $F_t$ to tension a pinch point, the positions of its neighbors can be calculated by using the Hooke's law \cite{09}:

\begin{equation*}
p_i^{t+1} = \alpha p^t_i + \delta (p_i^t - p_i^{t-1}) - \sum_{j \in \Delta_N^i - \Delta_C} \tau (p_j^t - p_i^t) + (F_t + g(t)),
\end{equation*}

\noindent
where $\alpha$ and $\delta$ represent time-constant parameters, $\tau$ represents a spring constant, and $g(t)$ denotes gravity. The vector of gravity $\overrightarrow{g(t)}$ belongs to the z-axis. For simplicity, we assume $\overrightarrow{F(t)}|_z = 0$. Let $A$ be a set of actions, we can define a tensioning policy $\pi_T$ as a mapping function from $L^t_\Sigma$ to probability distribution of $A$, \emph{i.e.}, $\pi^t_T: L^t_\Sigma \rightarrow P^t_A(.)$.

\subsection{Multi-Point Deep RL Tensioning Method}

\begin{figure}[!h]
\centering
\includegraphics[width=1.\linewidth]{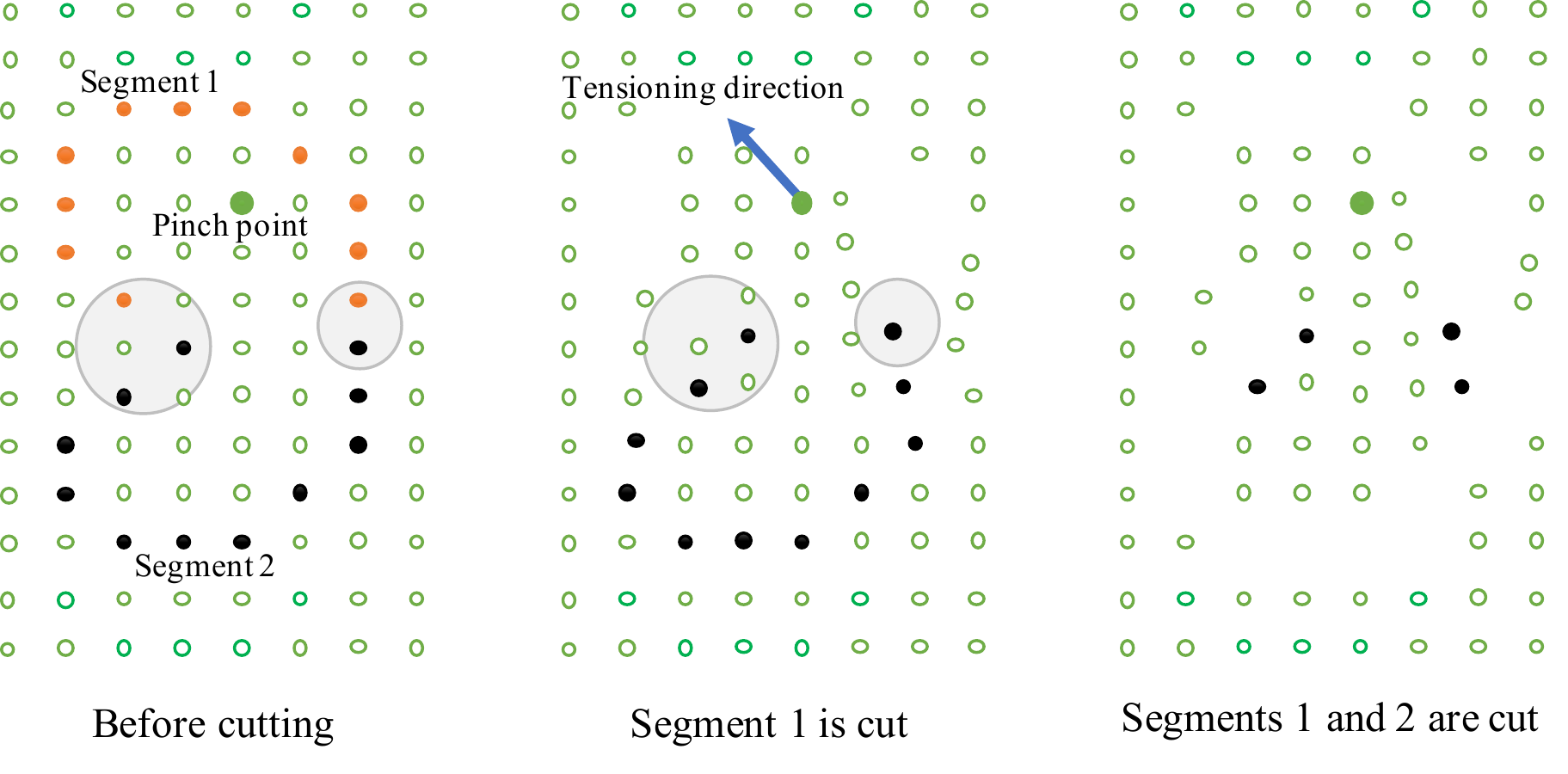}
\caption{Limitations of the use of one pinch point.}
\label{fig:3} 
\end{figure} 

A robot is normally limited by spatial and mechanical constraints. Therefore, it is intractable to use a scissor to cut a complicated contour without interruptions. One solution is to divide the contour into multiple segments and find the cutting order among these segments to minimize the damage to the material \cite{09}. However, the use of one pinch point makes it impossible to avoid any damage to the material, especially near the joint areas between segments. In Fig. \ref{fig:3}, for example, the contour is divided into two segments: segment 1 is illustrated by the orange dots and segment 2 is illustrated by the dark blue dots. The pinch point is represented by a solid green dot. The gray areas denote the joint areas between the segments. After segment 1 is completely cut, a tensioning force applied to the pinch point inadvertently causes the joint areas to distort, \emph{i.e.}, we start cutting segment 2 from an improper position. This drastically reduces the cutting accuracy. To overcome these obstacles, we add a fixed pinch point in each joint area to avoid distortion. This approach is feasible as it is done in the setup phase, which can be arranged before the cutting process. 

\begin{figure}[!t]
\centering
\includegraphics[width=1.\linewidth]{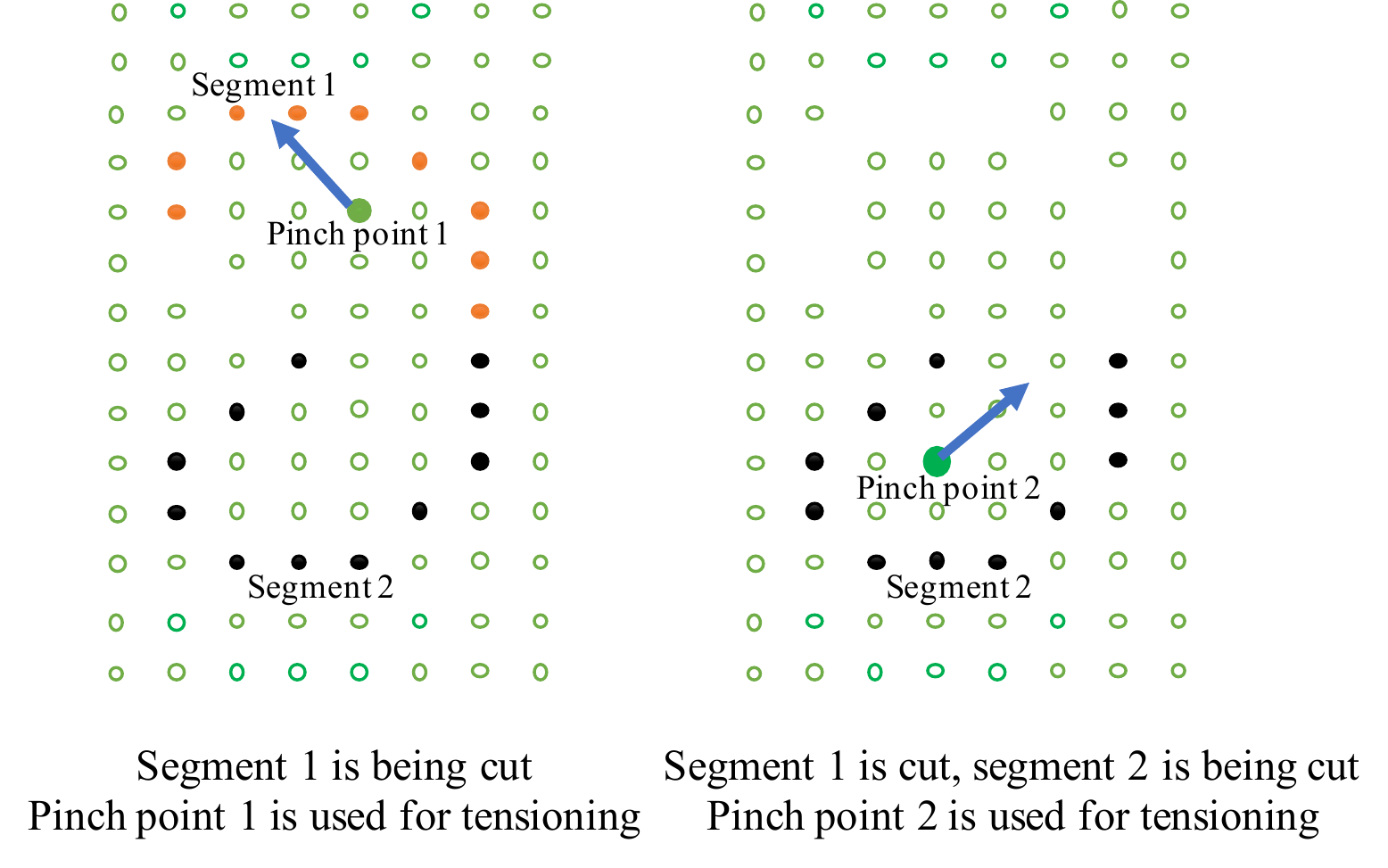}
\caption{The use of multiple pinch points for tensioning.}
\label{fig:4} 
\end{figure}

To further increase the efficiency by using pinch points, we proceed a \emph{divide and conquer} approach. In other words, if the contour is divided into $N$ segments, we find a set of $N$ different pinch points. Because we have one gripper, only one pinch point is used for tensioning. We also assume that the gripper while moving to a different pinch point does not affect the deformable sheet. This assumption is reasonable in surgical tasks where the deformable material is not too soft. In Fig. \ref{fig:4}, for example, we are cutting segment $i$ by using a pinch point $i$. After the segment $i$ is cut, the gripper selects a pinch point $j$ to start cutting segment $j$. This approach indicates that we need to find the best pinch point in each segment, which we call this process the local search. Previous work \cite{09} finds the best pinch point for all the segments, which is an intractable task. 

\begin{figure}[!b]
\centering
\includegraphics[width=1.\linewidth]{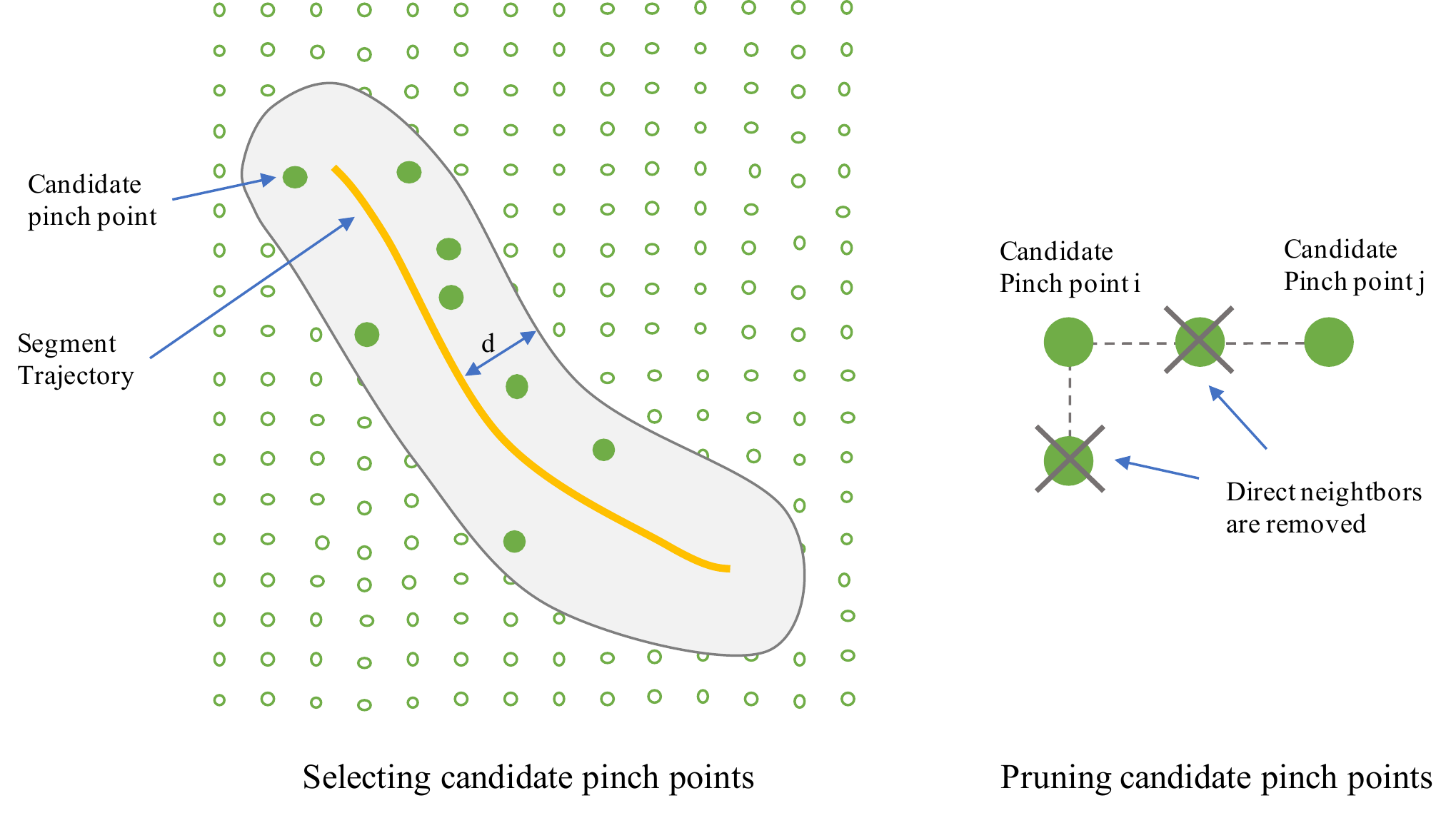}
\caption{Finding candidate pinch points of a segment.}
\label{fig:5} 
\end{figure}

A local search process for segment $i$ involves two steps: 1) finding a set of candidate pinch points for segment $i$ and 2) selecting the best pinch point among the candidates. Fig. \ref{fig:5} describes the process of finding a set of candidate pinch points for a specific segment $S$. Initially, we define a distance threshold $d > 0$, and then we find a set of candidate points around the segment based on $d$. Specifically, we select a point $p$ as a candidate point if it satisfies the following equation:

\begin{equation*}
\min_{i \in S} ||p - c_i|| < d,
\end{equation*}

\noindent
where $||p - c_i||$ denotes the distance between two points $p$ and $c_i$, and $c_i$ is a point in the segment $S$. After this step, we have a set of candidates $A$. We take $M$ candidates randomly from $A$ and put them in an empty set $B$. After that, we remove all candidates that are direct neighbors in $B$ to form a set $C$, as shown in Fig. \ref{fig:5}. The next step is to use the TRPO algorithm to create a tensioning policy for each candidate in $C$.

Fig. \ref{fig:6} summarizes the workflow of our Multi-point Deep RL Tensioning method (MDRLT) as follows:

\begin{figure}[!t]
\centering
\includegraphics[width=0.95\linewidth]{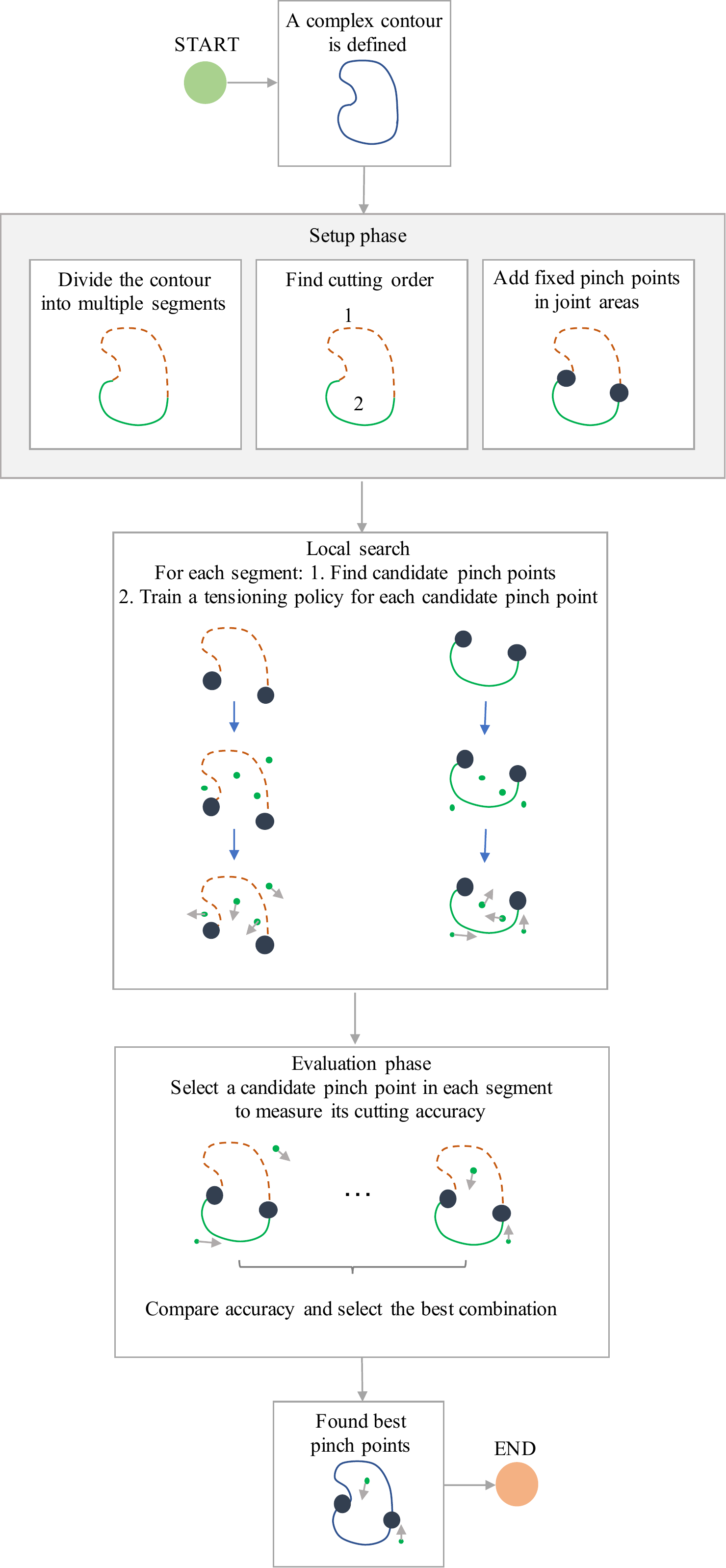}
\caption{A workflow of the MDRLT method.}
\label{fig:6} 
\end{figure} 

\begin{itemize}

\item \emph{Problem definition}: A complex contour is defined in the deformable material.

\item \emph{Setup phase}: This phase involves dividing the contour into multiple segments, finding cutting order between these segments, and finally adding fixed pinch points in joint areas.

\item \emph{Local search}: As described earlier, the goal of this phase is to find a set of candidate tensioning pinch points in each segment.

\item \emph{Evaluation phase}: This phase combines each candidate pinch point in each segment to evaluate the accuracy while cutting the whole contour.

\item \emph{Final phase}: The best combination of candidate pinch points is selected together with fixed pinch points in joint areas. This phase terminates our algorithm.

\end{itemize}

\section{Performance Evaluation}
\label{sec:4}

\subsection{Simulation settings}

In this section, we use the simulator described in \cite{09} with the following parameter settings: $g(t) = -2500$, $\alpha = 0.99$, $\delta = 0.008$, and $\tau = 1$. The threshold $d$ equals to 100. The maximum number of candidate pinch points in each segment is $M = 20$ if the number of segments equals to 2 and $M = 10$ if the number of segments is greater than 2. The algorithms to divide the contour into different segments are based on the mechanical constraints of the dVRK and developed in the simulator. To find the best cutting order among different segments, we use the exhaustive search to find the order that provides the highest accuracy. Each tensioning policy is trained with TRPO in 20 iterations, a batch size of 500, a step size of 0.01, and a discount factor of 1. We use the implementation of the TRPO algorithm as in \cite{33}. The cutting accuracy is also defined in the simulator, which is the symmetric difference between the ideal contour with the actual contour cut. The cutting reliability is measured by calculating the standard deviation while evaluating cutting accuracy. Finally, the simulator is significantly modified to support local search.

\subsection{Accuracy performance}

\begin{figure*}[!t]
\centering
\includegraphics[width=1.\linewidth]{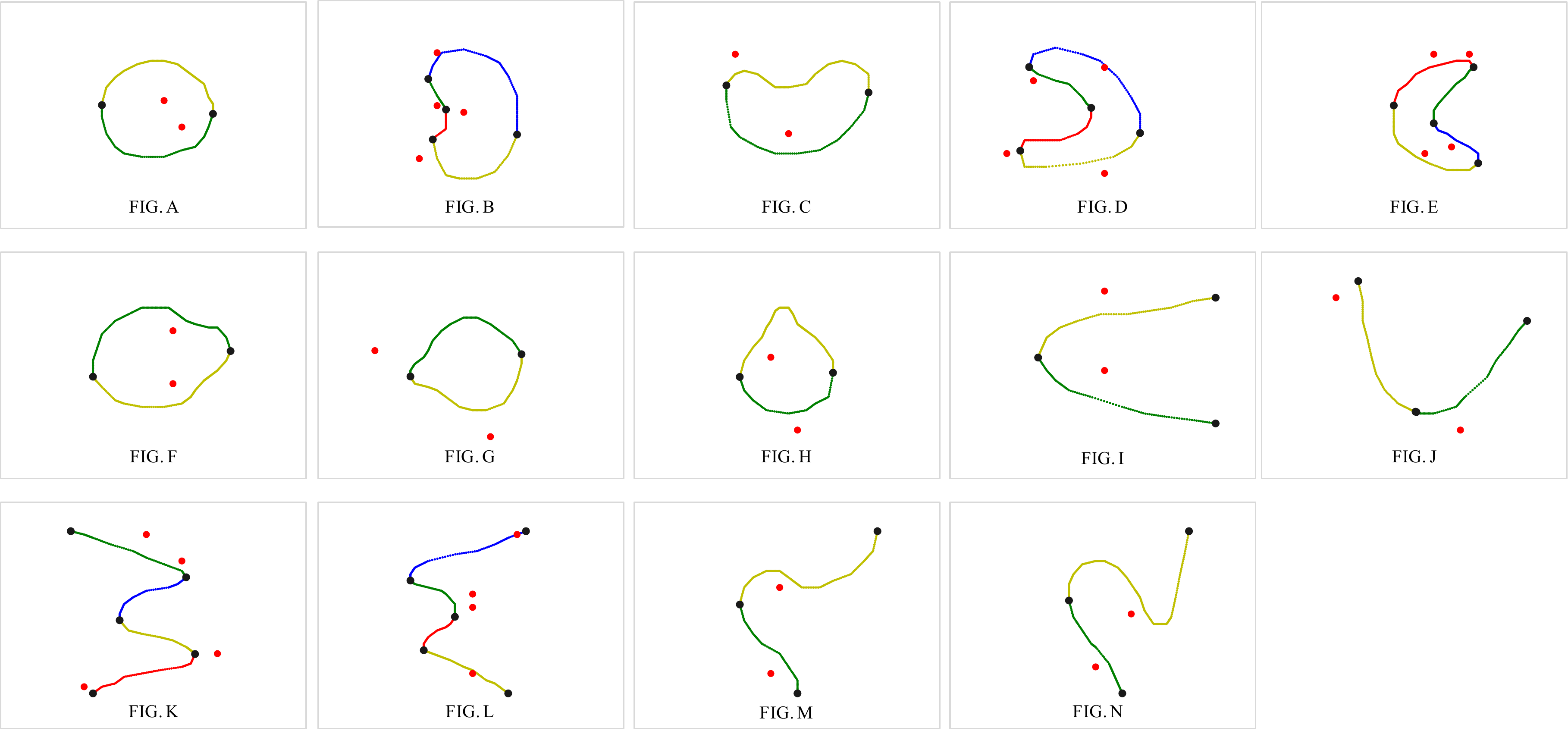}
\caption{A testbed of 14 different open and closed contours.}
\label{fig:7} 
\end{figure*} 

\begin{figure}[!h]
\centering
\includegraphics[width=1.\linewidth]{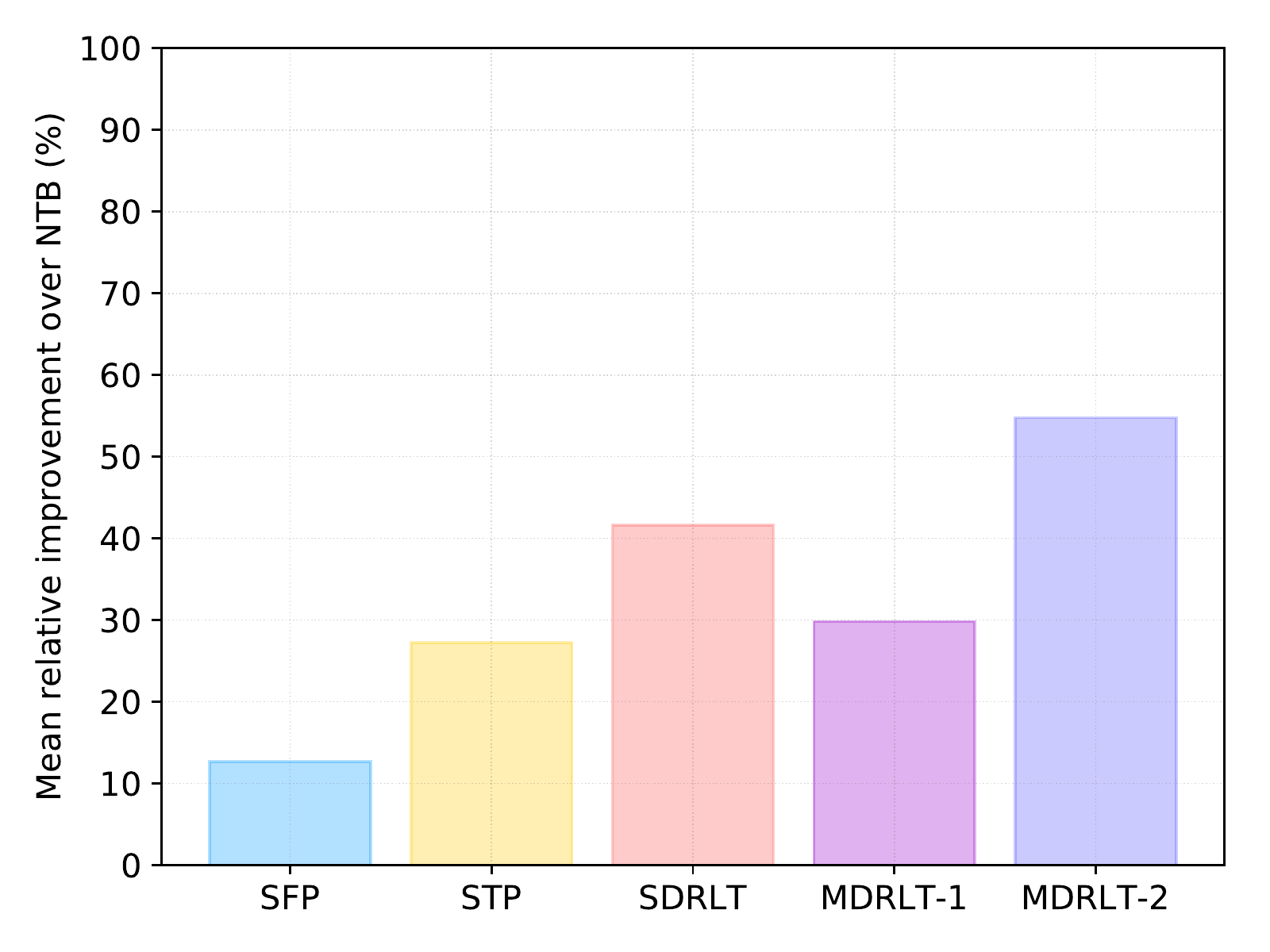}
\caption{The mean of relative percentage improvement over NTB of five algorithms.}
\label{fig:8} 
\end{figure} 

\begin{figure}[!h]
\centering
\includegraphics[width=1.\linewidth]{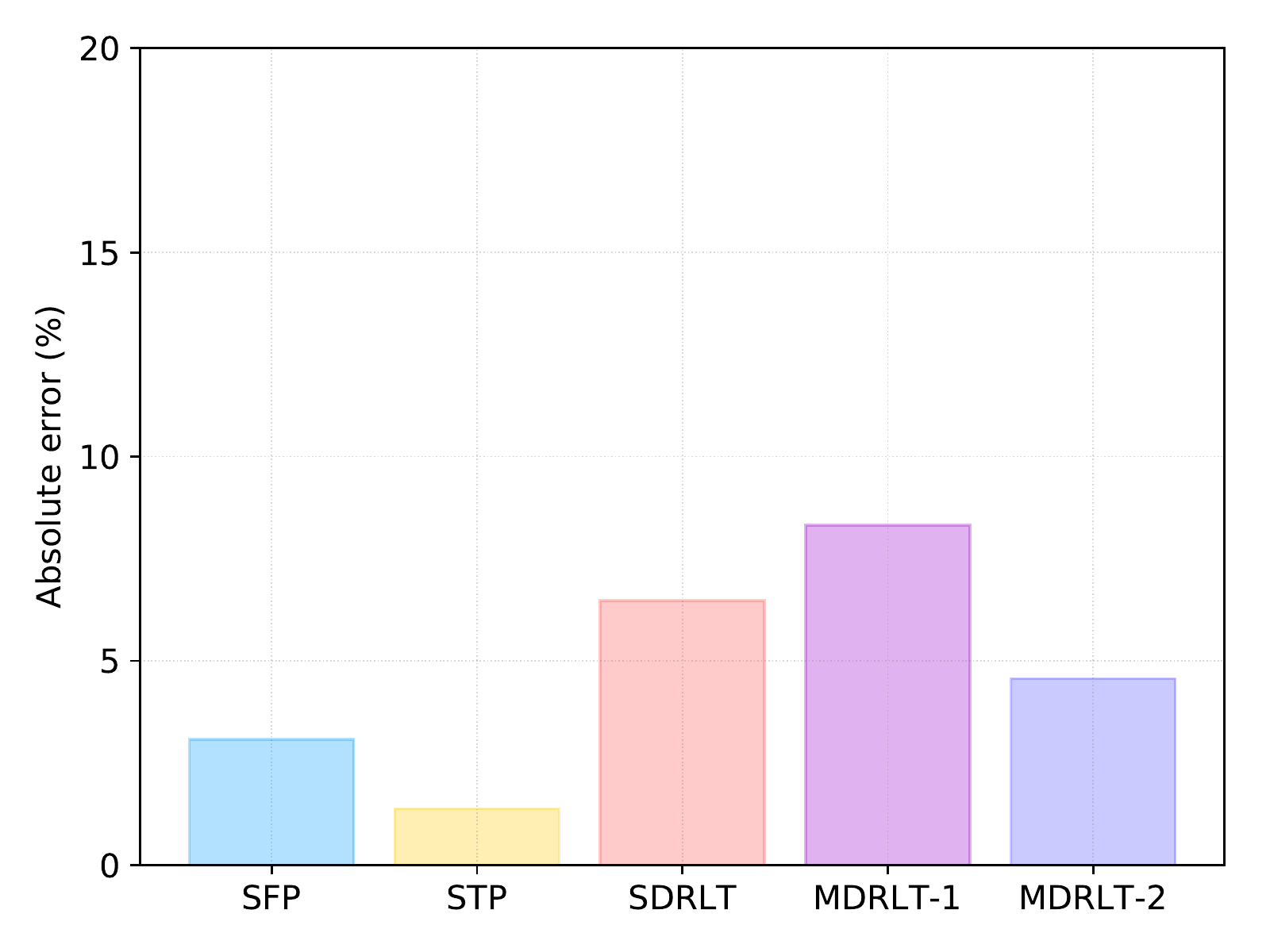}
\caption{The reliability comparison of five algorithms.}
\label{fig:9} 
\end{figure} 

\begin{table*}[!h]
  \caption{Raw values of the accuracy test using MDRLT-1 and MDRLT-2. Each figure is cut in 10 times to measure the symmetric difference between the ideal contour and the actual contour cut.}
  \label{table:1}
  \small
  \centering
  \begin{tabular}{lllllllllllll}
   \toprule
   \midrule
   \textbf{Contour} & \textbf{Algorithm} &\textbf{Eval 1}& \textbf{Eval 2}& \textbf{Eval 3}& \textbf{Eval 4}& \textbf{Eval 5}& \textbf{Eval 6}& \textbf{Eval 7}& \textbf{Eval 8}& \textbf{Eval 9}& \textbf{Eval 10} & \textbf{Mean}\\
   \midrule
   Figure A & MDRLT-1& 29& 31& 46& 34& 50& 31& 40& 38& 40& 39& 37.8\\\\
   		        & MDRLT-2& 24& 37& 35& 32& 33& 45& 37& 30& 37& 27& \textbf{33.7}\\
   \midrule
   Figure B & MDRLT-1& 33& 67& 55& 53& 50& 60& 42& 36& 33& 54&48.3\\\\
   		        & MDRLT-2& 27& 39& 42& 79& 27& 26& 23& 25& 38& 28&\textbf{35.4}\\
   \midrule
   Figure C & MDRLT-1& 36& 44& 44& 42& 47& 34& 50& 43& 45& 53&43.8\\\\
   		        & MDRLT-2& 39& 30& 26& 34& 26& 35& 33& 32& 39& 33&\textbf{32.7}\\
   \midrule
   Figure D & MDRLT-1& 144& 137& 130& 139& 136& 136& 137& 129& 129& 133&135\\\\
   		        & MDRLT-2& 40& 34& 42& 36& 45& 32& 36& 44& 39& 25&\textbf{37.3}\\
   \midrule
   Figure E & MDRLT-1& 65& 65& 59& 55& 64& 52& 51& 74& 62& 56&60.3\\\\
   		        & MDRLT-2& 38& 44& 34& 51& 34& 39& 35& 36& 36& 36&\textbf{38.3}\\
   \midrule
   Figure F & MDRLT-1& 58& 70& 49& 44& 42& 53& 53& 33& 37& 35&47.4\\\\
   		        & MDRLT-2& 36& 38& 34& 28& 35& 28& 25& 28& 40& 27&\textbf{31.9}\\
   \midrule
   Figure G & MDRLT-1& 38& 41& 36& 44& 45& 33& 48& 48& 38& 42&41.3\\\\
   		        & MDRLT-2& 19& 25& 22& 23& 22& 22& 26& 25& 29& 24&\textbf{23.7}\\
   \midrule
   Figure H & MDRLT-1& 43& 18& 16& 46& 39& 28& 25& 23& 28& 18&28.4\\\\
   		        & MDRLT-2& 17& 17& 20& 20& 13& 24& 17& 16& 17& 21&\textbf{18.2}\\
   \midrule
   Figure I & MDRLT-1& 78& 62& 73& 62& 51& 55& 51& 74& 76& 77&65.9\\\\
   		        & MDRLT-2& 37& 40& 42& 44& 38& 43& 39& 52& 39& 46&\textbf{42}\\
   \midrule
   Figure J & MDRLT-1& 19& 18& 13& 12& 20& 19& 21& 18& 19& 19 &\textbf{17.8}\\\\
   		        & MDRLT-2& 25& 26& 21& 22& 21& 21& 21& 21& 21& 21&22\\
   \midrule
   Figure K & MDRLT-1& 82& 104& 79& 87& 90& 89& 76& 78& 74&59&81.8\\\\
   		        & MDRLT-2& 48& 47& 39& 45& 56& 57& 49& 81& 42& 43&\textbf{50.7}\\
   \midrule
   Figure L & MDRLT-1& 35& 31& 37& 32& 41& 32& 35& 32& 33& 33&34.1\\\\
   		        & MDRLT-2& 14& 15& 18& 12& 12& 15& 15& 14& 15& 20&\textbf{15}\\
   \midrule
   Figure M & MDRLT-1& 21& 24& 20& 21& 21& 25& 23& 18& 17& 23&21.3\\\\
   		        & MDRLT-2& 14& 15& 14& 13& 19& 15& 14& 17& 17& 14&\textbf{15.2}\\
   \midrule
   Figure N & MDRLT-1& 38& 34& 102& 25& 34& 26& 28& 32& 18& 56&39.3\\\\
   		        & MDRLT-2& 23& 19& 20& 27& 18& 26& 26& 30& 23& 23&\textbf{23.5}\\
   \midrule
    \bottomrule
  \end{tabular}
\end{table*}

To compare the cutting accuracy between different algorithms, we select 14 complicated multi-segment contours described in \cite{09}, as shown in Fig. 7. The black dots represent the fixed pinch points that are used in the setup phase. The red dots represent the best tensioning pinch points found by the local search. We compare the cutting accuracy and reliability between six algorithms (the first four algorithms are based in \cite{09}):

\begin{itemize}
\item \emph{Non-Tensioned Baseline (NTB)}: We only use the scissor to cut the contour without using the gripper.

\item \emph{Single Fixed Pinch point without tensioning (SFP)}: A single fixed pinch point is used without tensioning.

\item \emph{Single Tensioning Pinch point (STP)}: A single tensioning pinch point is used.

\item \emph{SDRLT}: A single tensioning pinch point is used. We use the TRPO to find the tensioning policy for the pinch point.

\item \emph{MDRLT-1}: The proposed algorithm without using the fixed pinch points in joint areas.

\item \emph{MDRLT-2}: The proposed algorithm using both fixed pinch points and tensioning pinch points.

\end{itemize}

We evaluate the proposed algorithms (MDRLT-1 and MDRLT-2) in 10 simulated trials for each figure in the testbed. Table \ref{table:1} presents the raw values of the symmetric difference between the ideal contour and the actual contour cut in this evaluation. Fig. \ref{fig:8} shows the mean of relative percentage improvement in symmetric difference over the NTB method of five different algorithms. We see that the use of fixed pinch points during the setup phase determines the cutting accuracy. Therefore, MDRLT-1 is not better than SDRLT but MDRLT-2 significantly outperforms SDRLT, which is the state-of-the-art method in surgical pattern cutting. 

Finally, Fig. \ref{fig:9} shows the absolute error while evaluating the cutting accuracy in 10 trials. This metrics represents the reliability of the proposed methods. Among three algorithms using deep RL, MDRLT-2 provides the highest reliability as it has the lowest value of absolute error.

\section{Conclusion}
\label{sec:5}

This paper introduces an interesting multi-point approach based on deep reinforcement learning for the surgical soft tissue cutting task that is meaningful in both practical perspective and theoretical analysis. In the theoretical analysis, the paper benchmarks the accuracy of the use of multiple pinch points during the course of cutting, which is the first study according to our best knowledge. The study also concludes that the use of fixed pinch points in joint areas is the key to significantly outperform the state-of-the-art cutting method with respect to accuracy and reliability. 

The proposed approach becomes a normative workflow to ensure the safety in the surgical pattern cutting task. Moreover, it can be applied to a diversity of future research such as the use of multiple grippers or multiple scissors in surgical tasks, multi-layer pattern cutting in 3D space, or 3D multi-segment contours.

\end{document}